\documentclass[11pt,a4paper]{article}
\usepackage[hyperref]{emnlp2018}
\usepackage{times}
\usepackage{latexsym}

\usepackage{url}
\usepackage{multirow}
\usepackage{graphicx}
\usepackage{lipsum}
\usepackage{enumitem}
\usepackage{amsmath}
\usepackage{xcolor}
\usepackage{subcaption}
\usepackage{tabularx}
\usepackage{CJKutf8}
\usepackage{booktabs}
\usepackage{mathrsfs}
\usepackage{cleveref}
\usepackage{amsmath}
\usepackage{amssymb}
\usepackage{hhline}
\usepackage{algorithm}
\usepackage{algorithmic}
\usepackage{xparse}
\usepackage{caption}

\aclfinalcopy % Uncomment this line for the final submission

\usepackage{etoolbox}
\makeatletter
\patchcmd\@combinedblfloats{\box\@outputbox}{\unvbox\@outputbox}{}{%
   \errmessage{\noexpand\@combinedblfloats could not be patched}%
}%
\makeatother

\NewDocumentCommand{\heng}{ mO{} }{\textcolor{orange}{\textsuperscript{\textit{Heng}}\textsf{\textbf{\small[#1]}}}}

\NewDocumentCommand{\done}{ mO{} }{\textcolor{BlueGreen}{\textsuperscript{\textit{#1}}\textsf{\textbf{\small[Done]}}}}
\definecolor{BlueGreen}{rgb}{0.0, 0.87, 0.87}

\DeclareCaptionLabelFormat{andtable}{#1~#2  \&  \tablename~\thetable}
\title{Entity-aware Image Caption Generation}

\author{Di Lu$^{1}$, Spencer Whitehead$^{1}$, Lifu Huang$^{1}, $Heng Ji$^{1}$, Shih-Fu Chang$^{2}$ \\
  $^{1}$Computer Science Department, Rensselaer Polytechnic Institute \\
  {\tt \{lud2,whites5,huangl7,jih\}@rpi.edu}\\
  $^{2}$Computer Science Department, Columbia University\\
  {\tt sfchang@cs.columbia.edu}
    }

\date{}

\begin{document}
\maketitle
\begin{abstract}

Current image captioning approaches generate descriptions which lack specific information, such as named entities that are involved in the images. In this paper we propose a new task which aims to generate informative image captions, given images and hashtags as input. We propose a simple but effective approach to tackle this problem. We first train a convolutional neural networks - long short term memory networks (CNN-LSTM) model to generate a template caption based on the input image. Then we use a knowledge graph based collective inference algorithm to fill in the template with specific named entities retrieved via the hashtags. Experiments on a new benchmark dataset collected from Flickr show that our model generates news-style image descriptions with much richer information. 
Our model outperforms unimodal baselines significantly with various evaluation metrics.

\footnote{Datasets and programs: \url{https://github.com/dylandilu/Entity-aware-Image-Captioning}}

\end{abstract}

\section{Introduction}
\label{sec:intro}

As information regarding emergent situations disseminates through social media, the information is presented in a variety of data modalities ({\sl e.g.} text, images, and videos), with each modality providing a slightly different perspective.
Images have the capability to vividly represent events and entities, but without proper contextual information they become less meaningful and lose utility. While images may be accompanied by associated tags or other meta-data, which are inadequate to convey detailed events, many lack the descriptive text to provide such context. For example, there are 17,285 images on Flickr from the Women's March on January 21, 2017,\footnote{https://en.wikipedia.org/wiki/2017\_Women\%27s\_March} most of which contain only a few tags and lack any detailed text descriptions. The absence of context leaves individuals with no knowledge of details such as the purpose or location of the march.

Image captioning offers a viable method to directly provide images with the necessary contextual information through textual descriptions. Advances in image captioning~\cite{xu2015show,fang2015captions,karpathy2015deep,vinyals2015show} are effective in generating sentence-level descriptions. However, sentences generated by these approaches are usually generic descriptions of the visual content and ignore %any and all 
background information. Such generic descriptions do not suffice in emergent situations as they, essentially, mirror the information present in the images and do not provide detailed descriptions regarding events and entities present in, or related to, the images, which is imperative to understanding emergent situations. 
\begin{figure*}[ht]
\centering
\includegraphics[width=0.68\linewidth]{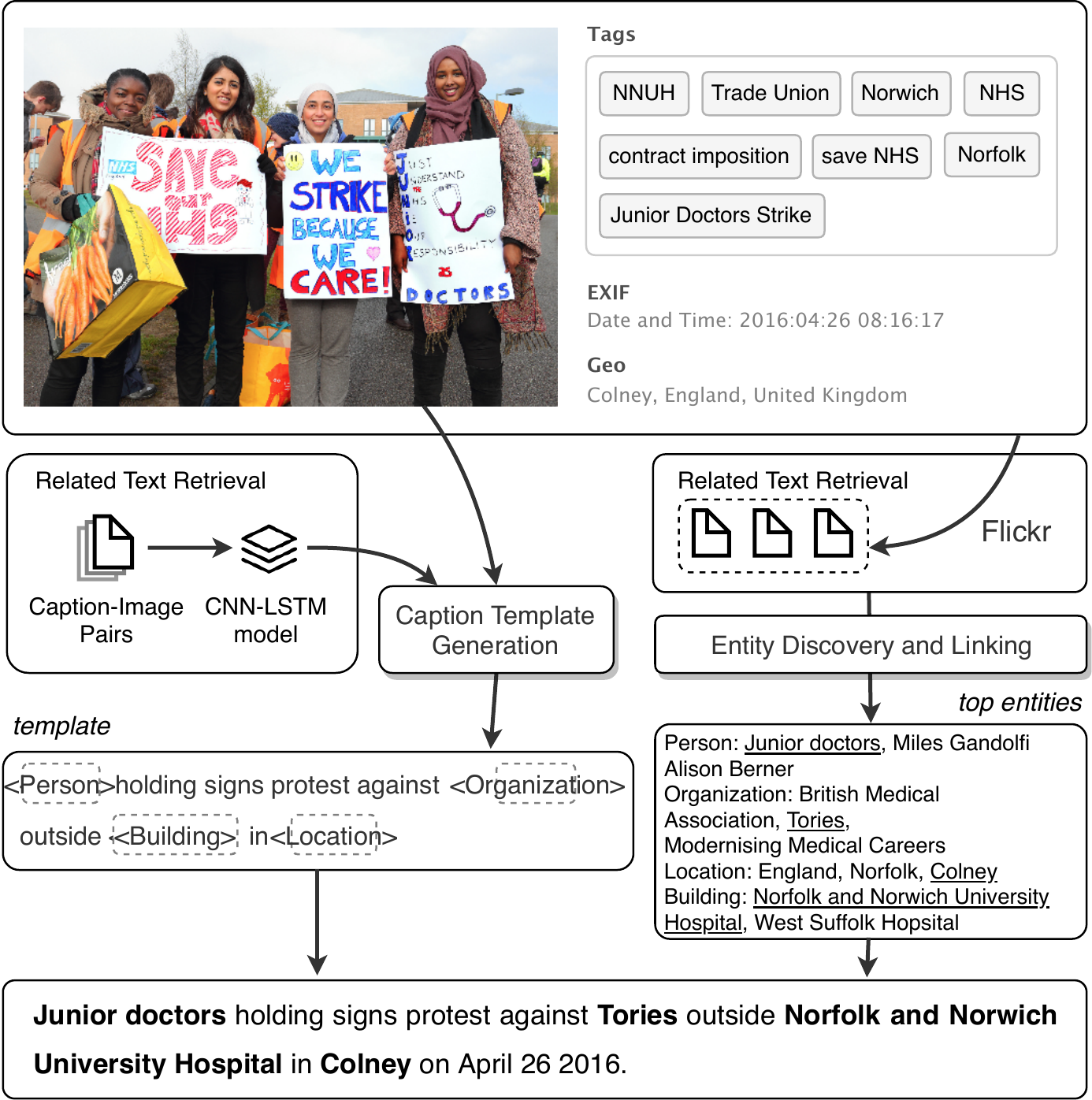}
\caption{The overall framework.}
\label{fig:overallframe}
\end{figure*}
For example, given the image in Figure~\ref{fig:overallframe}, the state-of-the-art automatically generated caption is `\textit{Several women hold signs in front of a building.}', which is lacking information regarding relevant entities ({\sl e.g.} `~\textit{Junior doctors}', `\textit{Tories}').

In our work, we propose an ambitious task: \textbf{entity-aware image caption generation}: automatically generate an image description that incorporates specific information such as named entities, relevant to the image, given limited text information, such as associated tags and meta-data ({\sl e.g.} time of photo and geo-location).
Our approach to this task generally follows three steps. First, instead of directly generating a sentence for an image, we generate a \textbf{\textit{template}} sentence with fillable slots by training an image captioning architecture on image-caption pairs, where we replace the entities from the captions with slot types indicating the type of entity that should be used to fill the slot. A \textbf{\textit{template}} for the example in Figure~\ref{fig:overallframe} is: \texttt{$<$Person$>$} holding signs protest against \texttt{$<$Organization$>$} outside \texttt{$<$Building$>$} in \texttt{$<$Location$>$}.

Second, given the associated tags of an image, we apply entity discovery and linking (EDL) methods to extract specific entities from previous posts that embed the same tags. Finally, we select appropriate candidates for each slot based upon the entity type and frequency. For example, we select the person name `\textit{Junior doctors}' to fill in the slot \texttt{$<$Person$>$} because it co-occurs frequently with other entities such as `\textit{Tories}' in the text related to the tags $\#NHS,\#Junior Doctors Strike$. This framework offers a distinct advantage in that it is very flexible, so more advanced captioning or EDL methods as well as other data can be used almost interchangeably within the framework.

To the best of our knowledge, we are the first to incorporate contextual information into image captioning without large-scale training data or topically related news articles and the generated image captions are closer to news captions.

\section{Approach Overview}
Figure~\ref{fig:overallframe} shows the overall framework of our proposed model. Given an image with associated tags and other meta-data, such as geographical tags and EXIF data,\footnote{EXIF data contains meta-data tags of photos such as date, time, and camera settings.} we first feed the image into a template caption generator to generate a sentence composed of context %vocabulary 
words, such as ``\textit{stand}'', and slots, such as \texttt{<person>}, to represent missing specific information like named entities (Section~\ref{sec:caption}). The template caption generator, which follows the encoder-decoder model~\cite{cho2014learning} with a CNN encoder and LSTM decoder~\cite{vinyals2015show}, is trained using news image-template caption pairs.

We then retrieve topically-related images from the Flickr database, which have the same tags as the input image. Next, we apply EDL algorithms to the image titles to extract entities and link them to external knowledge bases to retrieve their fine-grained entity types. Finally, for each slot generated by the template generator, we choose to fill the slot with the appropriate candidate based on entity type and frequency (Section~\ref{sec:filling}).

\section{Template Caption Generation}
\label{sec:caption}
Language models (LM) are widely used to generate text~\citep{wen2015semantically,tran2017nlg} and play a crucial role in most of the existing image captioning approaches~\citep{vinyals2015show,xu2015show}. These models, learned from large-scale corpora, are able to predict a probability distribution over a vocabulary. However, LM struggle to generate specific entities, which occur sparsely, if at all, within training corpora. Moreover, the desired entity-aware captions may contain information not directly present in the image alone. Unless the LM is trained or conditioned on data specific to the emergent situation of interest, the LM alone cannot generate a caption that incorporates the specific background information. We address this issue by only relying on the LM %language model 
to generate abstract slot tokens and connecting words or phrases, while slot filling is used to incorporate specific information. This approach allows the LM %language model 
to focus on generating words or phrases with higher probability, 
since each slot token effectively has a probability equal to the sum of all the specific entities that it represents, thereby circumventing the issue of generating lower probability or out-of-vocabulary (OOV) words.

In this section, we describe a novel method to train a model to automatically generate template captions with slots as `\textit{placeholders}' for specific background information. We first present the schemas which define the slot types (Section~\ref{sec:slots}) and the procedure to acquire training data for template generation (Section~\ref{sec:trainingdata}). Finally, we introduce the model for template caption generation (Section~\ref{sec:model}). 

\subsection{Template Caption Definition}
\label{sec:slots}

Named entities are the most specific information which cannot be easily learned by LM. Thus, in this work, we define slots as placeholders for entities with the same types. We use the fine grained entity types defined in DBpedia
\footnote{We use the sixth level entity types in Yago ontology (Wordnet types only).}
~\cite{auer2007dbpedia} to name the slots because these types are specific enough to differentiate between a wide range of entities and still general enough so that the slots have higher probabilities in the language model. For example, \texttt{Person} is further divided into \texttt{Athlete, Singer, Politician}, and so on.
Therefore, a template caption like `\texttt{Athlete} celebrates after scoring.' can be generated by the language model through leveraging image features, where the slot \texttt{Athlete} means a sports player (e.g., Cristiano Ronaldo, Lionel Messi).

\subsection{Acquisition of Image-Template Caption Pairs}
\label{sec:trainingdata}

\begin{figure}[ht]
\centering
\includegraphics[width=0.9\linewidth]{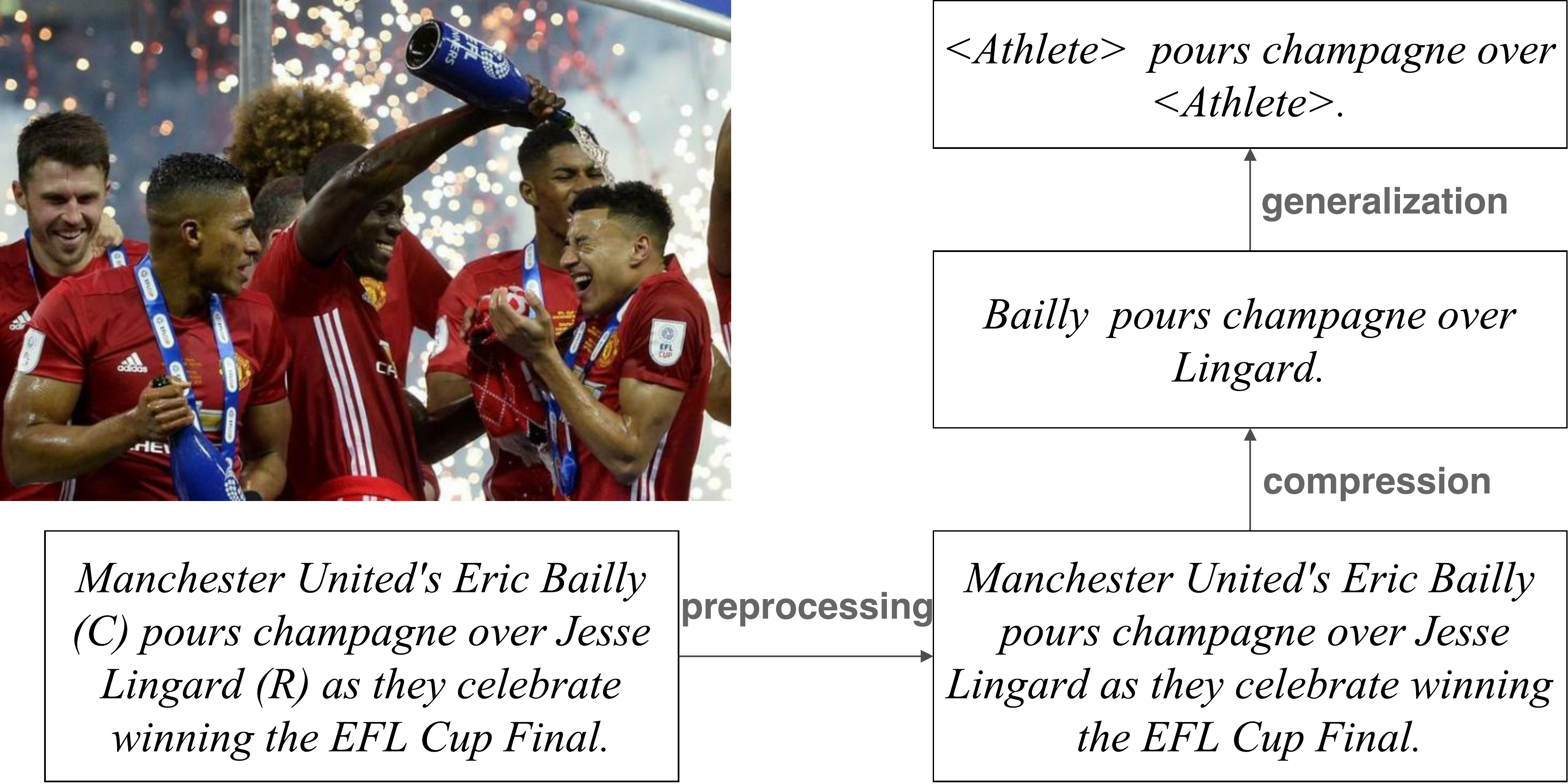}
\caption{Procedure of Converting News Captions into Templates.}
\label{fig:process}
\end{figure}

High quality training data is crucial to train a template caption generator. However, the image-caption datasets used in previous work, such as Microsoft Common Objects in Context (MS COCO)~\cite{lin2014microsoft} and Flickr30K~\cite{rashtchian2010collecting}, are not suitable for this task because they are designed for non-specific caption generation and do not contain detailed, specific information such as named entities. Further, manual creation of captions is expensive. In this work, we utilize news image-caption pairs, which are well written and can be easily collected.
We use the example in Figure~\ref{fig:process} to describe our procedure to convert image-caption to image-template caption pairs: \textbf{preprocessing}, \textbf{compression}, \textbf{generalization}.

\textbf{Preprocessing:} %Based on the criteria we posit in Section~\ref{sec:intro}, we 
We first apply the following pre-processing steps: (1) remove words in parentheses, such as \textit{`(C)'} and  \textit{`(R)'} in Figure~\ref{fig:process}, because they usually represent auxiliary information and are not aligned with visual concepts in images; (2) if a caption includes more than one sentence, we choose the longer one. Based on our observation, shorter sentences usually play the role of background introduction, which are not aligned with the key content in images; (3) remove captions with less than 10 tokens because they tend to be not informative enough. The average length of the news image captions is 37 tokens.

\textbf{Compression:} The goal of compression is to make news captions short and aligned with images as much as possible by keeping information related to objects and concepts in the images, which are usually subjects, verbs and objects in sentences.
% According to our observation, they usually appear as clause and tell background knowledge. 
In this paper, we propose a simple but efficient compression method based on dependency parsing. We do not use other complicated compressions~\cite{kuznetsova2014treetalk} because our simple method achieves comparative results on image caption dataset. We first apply the Stanford dependency parser~\cite{de2008stanford} on preprocessed captions. 
% \done{done}\heng{replace the following paragraph with a table that summarizes all heuristic rules you used for filtering; and replace "the words in red" with something more clear and formal}
Then, we traverse the parse tree from the root ({\sl e.g.}\textit{`pours'}) via $<$\texttt{governor}, \texttt{grammatical relations}, \texttt{dependent}$>$ triples using breadth-first search. We decide to keep a dependent or not based on its grammatical relation with the governor. 
Based on our observations, among the 50 grammatical relations in the Stanford dependency parser, we keep the dependents that have the following grammatical relations with their governors: \textit{nsubj, obj, iobj, dobj, acomp, det, neg, nsubjpass, pobj, predet, prep, prt, vmod, nmod, cc}.
 
\textbf{Generalization:} The last step for preparing training data is to extract entities from captions and replace them with the slot types we defined in Section~\ref{sec:slots}.
We apply Stanford CoreNLP name tagger~\cite{manning2014stanford} to the captions to extract entity mentions of the following types: \texttt{Person}, \texttt{Location}, \texttt{Organization}, and \texttt{Miscellaneous}.
Next, we use an English Entity Linking algorithm~\cite{Pan2015} to link the entity mentions to DBpedia and retrieve their fine-grained types.\footnote{This yields a total of 95 types after manually cleaning.} We choose the higher level type if there are multiple fine-grained types for a name.
For example, the entity types of \textit{Manchester United}, \textit{Eric Bailly}, and \textit{Jesse Lingard} are \textit{SoccerTeam}, \textit{Athlete}, and \textit{Athlete}, respectively.
For entity mentions that cannot be linked to DBpedia, we use their coarse-grained entity types, which are the outputs of name tagger.

Finally, we replace the entities in the compressed captions with their corresponding slots:
\begin{description}
\item \textbf{Generalized Template}: $<$\texttt{Athlete}$>$ pours champagne over $<$\texttt{Athlete}$>$.
\end{description}

\subsection{Generation Model}
\label{sec:model}

\begin{figure}[ht]
\centering
\includegraphics[width=0.8\linewidth]{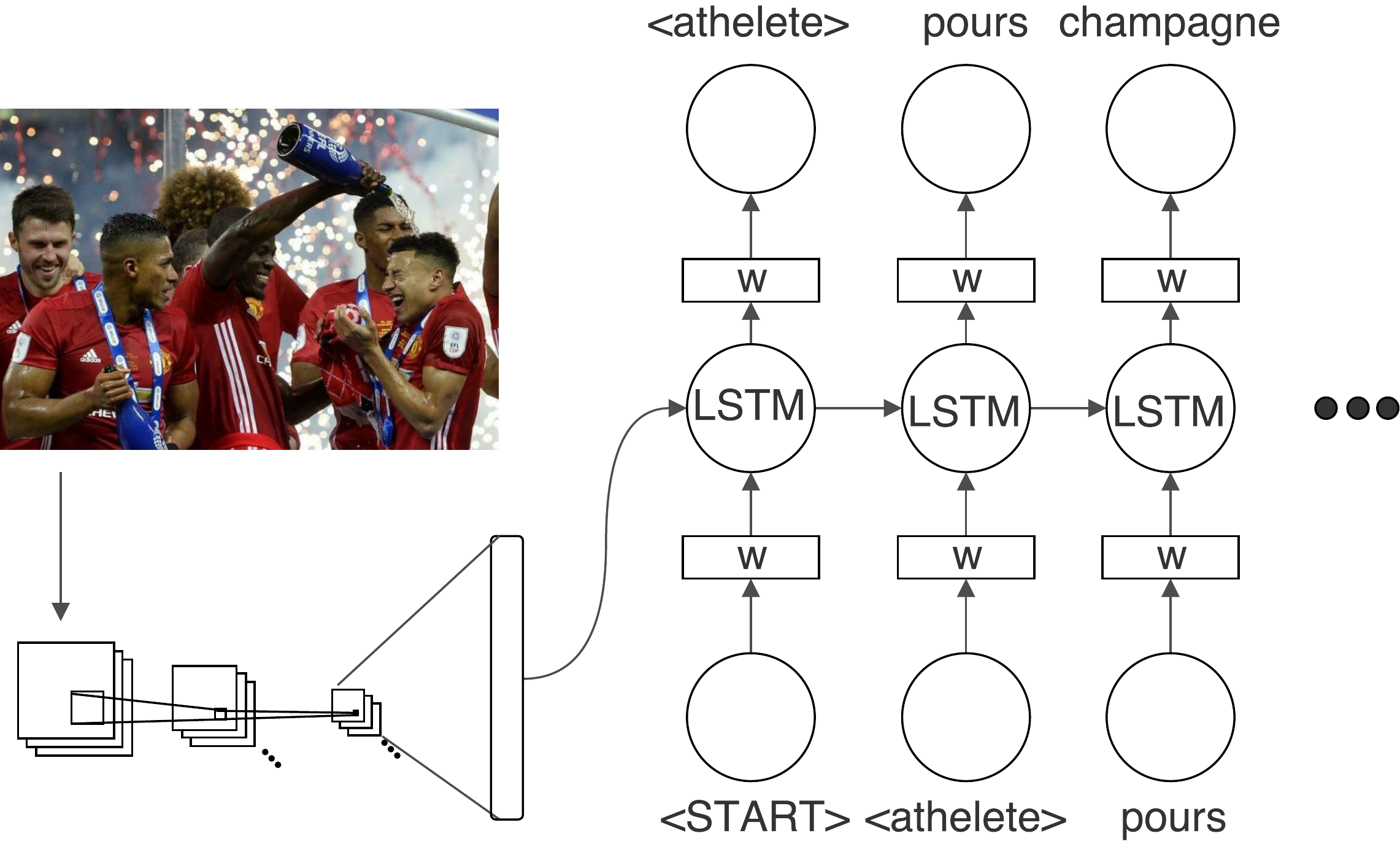}
\caption{LSTM language generator.}
% (a anti-$<politician>$ protest in $<loc>$, $<time>$.)}
\label{fig:lstm}
\end{figure}

Using the template caption and image pairs \textit{(S, I}) as training data, we regard the template caption generation as a regular image captioning task. Thus, we adapt the encoder-decoder architecture which is successful in the image captioning task~\cite{vinyals2015show,xu2015show}. Our model (Figure~\ref{fig:lstm}) is most similar to the one proposed in~\cite{vinyals2015show}. Note, other captioning methods may easily be used instead.

\textbf{Encoder:} Similar to previous work~\cite{vinyals2015show,xu2015show,karpathy2015deep,venugopalan2017captioning}, we encode images into representations using a ResNet~\cite{he2016deep} model pre-trained on the ImageNet dataset~\citep{imagenet_cvpr09} and use the outputs before the last fully-connected layer.

\textbf{Decoder:} We employ a Long Short Term Memory (LSTM)~\cite{hochreiter1997long} based language model to decode image representations into template captions. We provide the LSTM the image representation, $I$, as the initial hidden state.
%
% The LSTM is a special version of the RNN that can model long-range dependencies in a sequence as well as alleviate vanishing and exploding gradient problem. 
% The behavior of an LSTM cell is controlled by gates. 
At the $t^{\text{th}}$ step, the model predicts the probabilities of words/slots, $y_t$, based on the word/slot generated at last time step, $y_{t-1}$, as well as the hidden state, $s_t$.

\begin{figure*}[ht]
\centering
\includegraphics[width=0.8\linewidth]{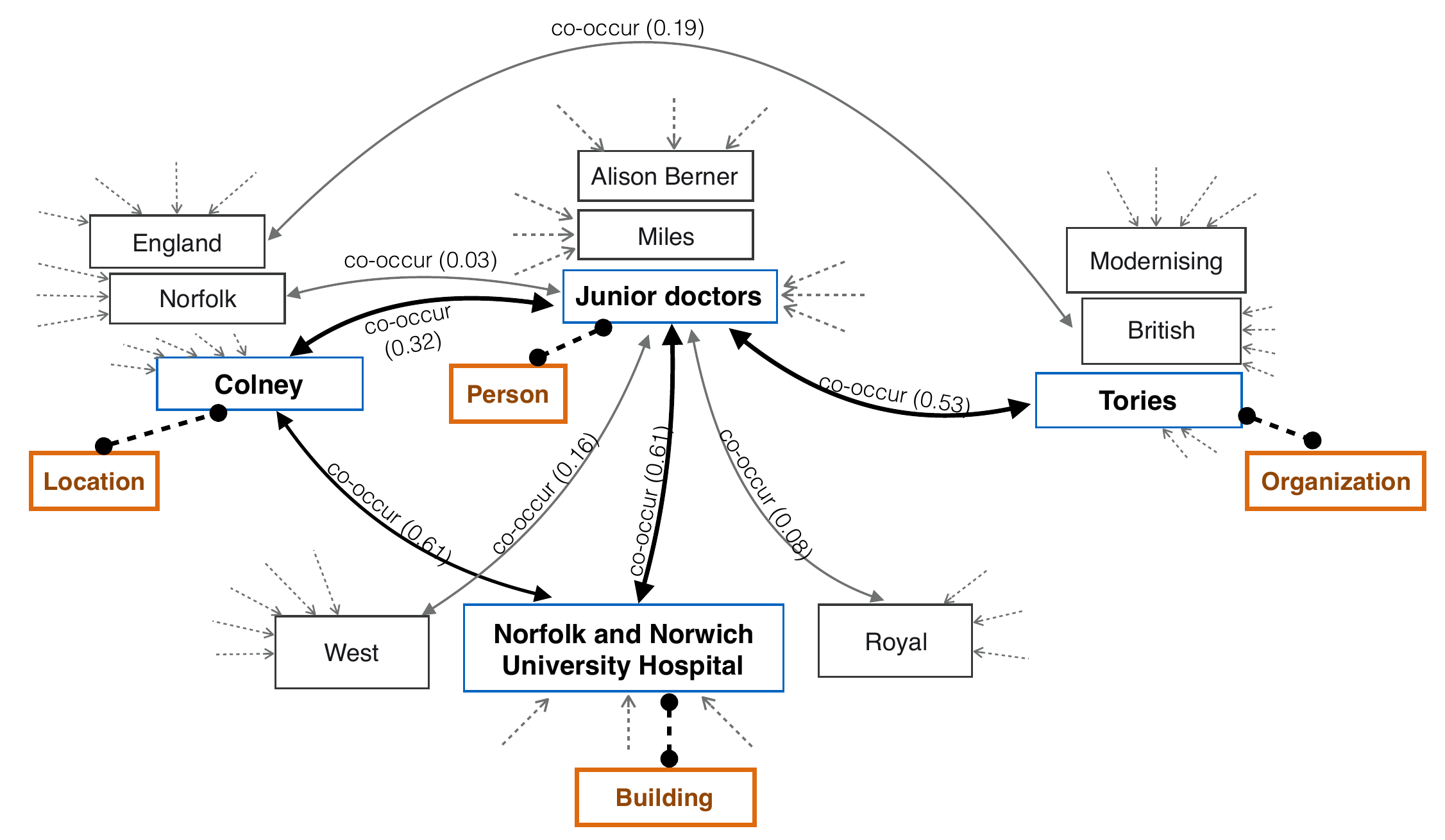}
\caption{Knowledge graph construction.}
\label{fig:kgmethod}
\end{figure*}
\section{Template Caption Entity Population}
\label{sec:filling}
With the generated template captions, our next step is to fill in the slots with the appropriate specific entities to make the caption complete and entity-aware.
% For example if a caption template contains a slot of $<$\texttt{club}$>$, and there's an associated tag of \texttt{\#lakers}, then we should fill \texttt{LA Lakers} into this slot.
%
In this section, we expand our method to extract candidate entities from contextual information ({\sl i.e.}, images in Flickr with the same tags). Once we extract candidate entities, we apply the Quantified Collective Validation (QCV) algorithm~\cite{wang2015}, which constructs a number of candidate graphs and performs collective validation on those candidate graphs to choose the appropriate entities for the slots in the template caption. 
\subsection{Candidate Entity Retrieval}

Users on social media typically post images with tags that are event-related (e.g. \#occupywallstreet), entity-related (e.g. \#lakers), or topic-related (e.g. \#basketball). On our Flickr testing dataset, the average number of tags associated with an image is 11.5 and posts with the same tags likely share common entities.
% \heng{provide concrete numbers about the percentage of images that have these associated tags} 
Therefore, given an image and its tags, we retrieve images from Flickr with the same tags by a window size of seven-day based on taken date of the photo, and then utilize the textual information accompanying the retrieved images as context. We filter out the high frequency hashtags($>200$ in testing dataset). Because some common tags, such as ‘\#concert’, appear in lots of posts related to different concerts.

Given the related text, we apply EDL algorithms~\cite{Pan2015} to extract named entities and link them to DBpedia to obtain their entity types. For each entity type, we rank the candidates based on their frequency in the context and only keep the top 5 candidate entities.

\subsection{Quantified Collective Validation}
Each slot in the template must be filled with an entity from its corresponding candidate entities. We can regard this step as an entity linking problem, whose goal is to choose an entity from several candidates given an entity mention. We utilize the QCV algorithm to construct a number of candidate graphs for a given set of slots~\cite{wang2015}, where each combination of candidate entities substituted into the slots yields a different graph (Figure~\ref{fig:kgmethod}). For each candidate combination graph, $G_c^i$, we compute the edge weights between each pair of candidates in the graph as
% \heng{give detailed formula} 
\begin{equation}
\label{eq:edge}
H_{r}=\frac{f_{c_hc_t}}{max(f_{c_h},f_{c_t})}
\end{equation}
where $r \in E(G_c^i)$ is an edge in $G_c^i$, $c_h$ and $c_t$ are the head candidate and tail candidate of the edge, $f_{c_hc_t}$ is the co-occurrence frequency of the pair of candidates, and $f_{c_h}$ and $f_{c_t}$ are the individual frequencies of head candidate and tail candidate, respectively. For example, in Figure~\ref{fig:kgmethod}, \textit{Colney} (\texttt{Location}) and \textit{Junior doctors} (\texttt{Person}) co-occur frequently, therefore the edge between them has a larger weight. 

We compute the summed edge weight, $\omega(G_c^i)$, for each $G_c^i$ by
% For each candidate graph, $G_c^i$, we use it's summed edge weight $\omega(G_c^i)$ as the ranking metric:
\begin{equation}
\omega(G_c^i)=\sum_{r \in E(G_c^i)} H_{r}
\end{equation}
and select the combination of candidates with the largest $\omega(G_c^i)$ to fill in the set of slots.

As a result of this process, given the template: `\texttt{$<$Person$>$} holding signs protest against \texttt{$<$Organization$>$} outside \texttt{$<$Building$>$} in \texttt{$<$Location$>$}.', we obtain an entity-aware caption: `\textit{\textbf{Junior doctors} holding signs protest against \textbf{Tories} outside \textbf{Norfolk and Norwich University Hospital} in \textbf{Colney}}'.

\subsection{Post-processing}
Some images in Flickr have EXIF data which gives the date that an image is taken. We convert this information into the format such as `\textit{\textbf{April 26 2016}}' and add it to the generated captions as post-processing, by which we obtain the complete caption: `\textit{Junior doctors holding signs protest against Tories outside Norfolk and Norwich University Hospital in Colney on \textbf{April 26 2016}.}'. We leave the generated caption without adding date information if it is not available. For those slots that cannot be filled by names, we use general words to replace them, such as using the word \textit{`Athlete'} to replace the slot \texttt{Athlete}.

% % \input{4visual}
% \input{5generation}
\section{Experiments}

\subsection{Data}
We require images with well-aligned, news-style captions for training. However, we want to test our model on real social media data and it is difficult to collect these informative captions for social media data. Therefore, we acquire training data from news outlets and testing data from social media. We select two different topics, social events and sports events, as our case studies. 
\begin{table}[ht!]
\centering
\small
\begin{tabular}
{lccc}
% {p{0.25\linewidth}|p{0.3\linewidth}p{0.14\linewidth}p{0.14\linewidth}p{0.1\linewidth}}
\toprule
{}&{Train}&{Dev}&{Test}\\
\midrule
{Number of Images}&{29,390}&{4,306}&{3,688}\\
{Number of Tokens}&{1,086,350}&{161,281}&{136,784}\\
{Social Event}&{6,998}&{976}&{872}\\
{Sports Event}&{22,392}&{3,330}&{2,816}\\
\midrule
{Person}&{32,878}&{4,539}&{4,156}\\
{Location}&{42,786}&{5,657}&{5,432}\\
{Organization}&{17,290}&{2,370}&{2,124}\\
{Miscellaneous}&{8,398}&{1,103}&{1,020}\\
% {}&\multicolumn{3}{c}{BLEU}&\multicolumn{3}{c}{METEOR}\\
\bottomrule
\end{tabular}
\caption{Statistics of datasets for template generation.}
\label{tab:reuters}
\end{table}

\textbf{Template Generator Training and Testing}. To train the template caption generator, we collect 43,586 image-caption pairs from Reuters\footnote{https://www.reuters.com/}, using topically-related keywords\footnote{Social Events: concert, festival, parade, protest, ceremony; Sports Events: Soccer, Basketball, Soccer, Baseball, Ice Hockey } as queries. We do not use existing image caption datasets, such as MSCOCO~\cite{lin2014microsoft}, because they do not contain many named entities. After the compression and generalization procedures (Section~\ref{sec:trainingdata}) we keep 37,384 images and split them into train, development, and test sets. Table~\ref{tab:reuters} shows the statistics of the datasets.

\textbf{Entity-aware Caption Testing}. Since news images do not have associated tags, for the purpose of testing our model in a real social media setting, we use images from Flickr for our caption evaluation\footnote{https://www.flickr.com/}, which is an image-centric, representative social media platform. 
%We chose two topics (demonstration and sports) in this paper, and 
We use the same keywords as for template generator training to retrieve multi-modal data  with Creative Commons license\footnote{https://creativecommons.org/licenses/}, for social and sports events. 
We choose the images that already have news-style descriptions from users and manually confirm they are well-aligned.
In total, we collect 2,594 images for evaluation.
% In total, we collect 503 images as evaluation set.
% number of entities in training data and testing data. overlapping and novel ones
For each image, we also obtain the tags (30,148 totally) and meta-data, such as EXIF and geotag data, when they are available. 

\subsection{Models for Comparison}
We compare our entity-aware model with the following baselines:

\noindent \textbf{CNN-RNN-raw}. We use the model proposed by~\citet{vinyals2015show} to train an image captioning model on the raw news image-caption pairs, and apply to Flickr testing data directly.

\noindent \textbf{CNN-RNN-compressed}. We use the model proposed by~\citet{vinyals2015show} to train a model on the compressed news image-caption pairs.

\noindent \textbf{Text-summarization}. We apply SumBasic summarization algorithms~\cite{vanderwende2007beyond}, that is a summarization for multiple documents based on frequency of word and semantic content units, to text documents retrieved by hashtag.

\noindent \textbf{Entity-aware}. We apply trained template generator on Flickr testing data, and then fill in the slots with extracted background information.

%knowledge.

\subsection{Evaluation Metrics}

We use three standard image captioning evaluation metrics, BLEU~\cite{papineni2002bleu} and METEOR~\cite{denkowski2014meteor}, REOUGE~\cite{lin2004rouge} and CIDEr~\cite{vedantam2015cider}, to evaluate the quality of both the generated templates and generated captions. BLEU is a metric based on correlations at the sentence level.
% , scoring generated sentences based upon the number of overlapping $n$-grams. We also incorporate 
METEOR is a metric with recall weighted higher than precision and it takes into account stemming as well as synonym matching. %, as they are traditional used in image captioning tasks. 
ROUGE is proposed for evaluation of summarization and relies highly on recall.
CIDEr metric downweights the n-grams common in all image captions, which are similar to tf-idf.
Since the goal of this task is to generate entity-aware descriptions, we also measure the entity F1 scores for the final captions, where we do fuzzy matching to manually count the overlapped entities between the system output and reference captions. Besides, we do human evaluation with a score in range of 0 to 3 using the criteria as follows: Score 0: generated caption is not related to the ground-truth; Score 1: generated caption is topically-related to the ground-truth, but has obvious errors; Score 2: generated caption is topically-related to the ground-truth, and has overlapped named entities; Score 3: generated caption well describes the image.

\subsection{Template Evaluation}
\begin{figure}[!ht]
\small
\centering
\includegraphics[width=0.5\linewidth]{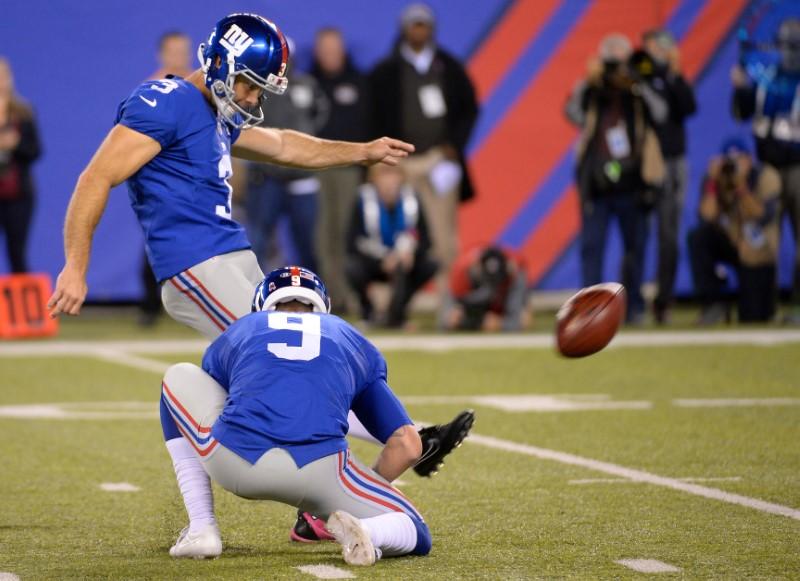}
\centering
\begin{tabular}{p{7cm}}
\toprule
\textbf{Raw}: new york giants kicker josh brown kicks a field goal during the first quarter against the san francisco 49ers at metlife stadium.\\
\midrule
\textbf{Generated Coarse Template}: $<$\texttt{Person}$>$ runs with ball against $<$\texttt{Location}$>$ $<$\texttt{Location}$>$ in the first half of their $<$\texttt{Miscellaneous}$>$ football game in $<$\texttt{Location}$>$\\
\midrule
\textbf{Generated Fine Template}: $<$\texttt{Footballteam}$>$ kicker $<$\texttt{Player}$>$ kicks a field goal out of the game against the $<$\texttt{Footballteam}$>$ at $<$\texttt{Organization}$>$ stadium\\
\bottomrule
\end{tabular}
% \captionsetup{labelformat=parens}
\caption{Example of generated template.}
\label{fig:template_ex}
\end{figure}

\begin{table*}[ht!]
\centering
\small
\scalebox{1}{
\begin{tabular}
{lcccccccc}
\toprule
{Approach}&{Vocabulary}&{BLEU-1}&{BLEU-2}&{BLEU-3}&{BLEU-4}&{METEOR}&{ROUGE}&{CIDEr}\\
\midrule
{Raw-caption}&{10,979}&{ 15.1}&{ 11.7}&{ 9.9}&{8.8}&{8.8}&{24.2}&{34.7}\\
{Coarse Template}&{3,533}&\textbf{46.7}&\textbf{36.1}&\textbf{29.8}&{\textbf{25.7}}&\textbf{22.4}&\textbf{43.5}&{161.6}\\
{Fine Template}&{3,642}&{43.0}&{33.4}&{27.8}&{24.3}&{20.3}&{39.8}&\textbf{165.3}\\
\bottomrule
\end{tabular}
}
\caption{Comparison of Template Generator with coarse/fine-grained entity type. Coarse Template is generalized by coarse-grained entity type (name tagger) and fine template is generalized by fine-grained entity type (EDL).}
\label{tab:template_results}
\end{table*}
Table~\ref{tab:template_results} shows the performances of template generator based on coarse-grained and fine-grained type respectively, and Figure~\ref{fig:template_ex} shows an example of the template generated. 
Coarse templates are the ones after we replace names with these coarse-grained types.
Entity Linking classifies names into more fine-grained types, so the corresponding templates are fine templates.
The generalization method of replacing the named entities with entity types can reduce the vocabulary size significantly, which reduces the impact of sparse named entities in training data. The template generation achieves close performance with state-of-the-art generic image captioning on MSCOCO dataset~\cite{xu2015show}. The template generator based on coarse-grained entity type outperforms the one based on fine-grained entity type for two reasons: (1) fine template relies on EDL, and incorrect linkings import noise; (2) named entities usually has multiple types, but we only choose one during generalization. Thus the caption, `\textit{Bob Dylan performs at the Wiltern Theatre in Los Angeles}', is generalized into `\textit{$<$\texttt{Writer}$>$ performs at the $<$\texttt{Theater}$>$ in $<$\texttt{Loaction}$>$}', but the correct type for \textit{Bob Dylan} in this context should be \texttt{Artist}.

\begin{figure*}[!ht]
\centering
\includegraphics[width=0.7\linewidth]{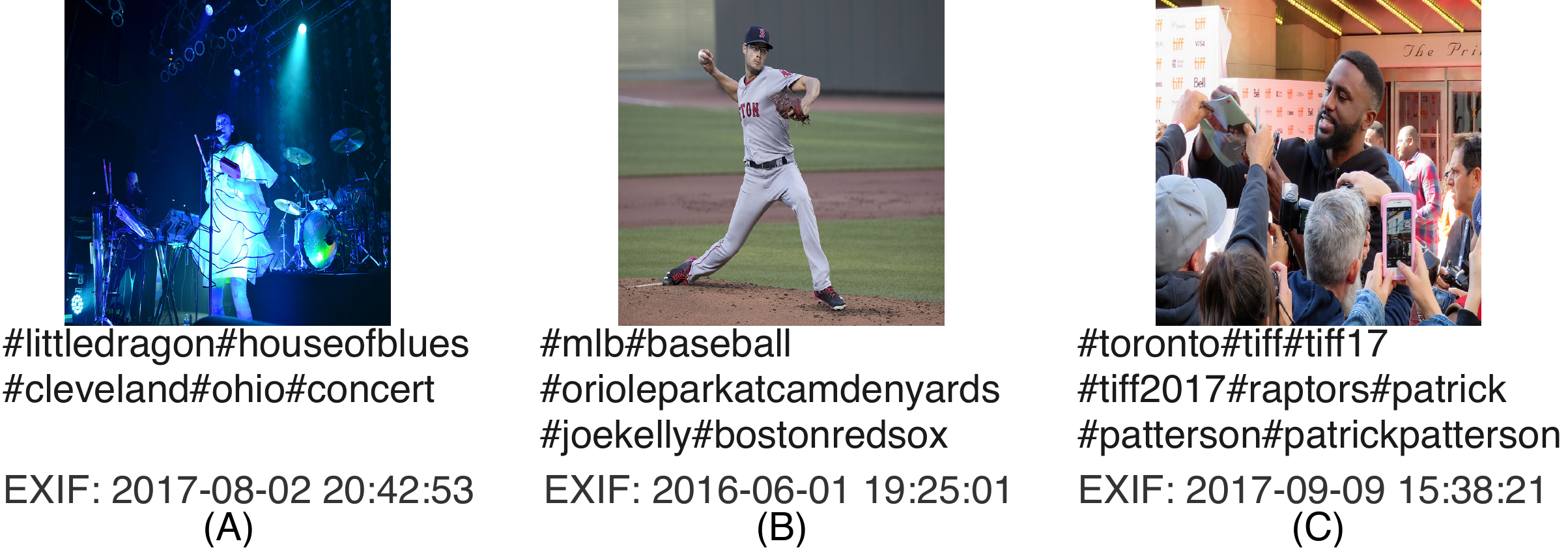}
\centering
\small
\begin{tabular}{c|cm{0.7\linewidth}}
\toprule
{}&{\textbf{Model}}&{\textbf{Caption}}\\
\midrule
\multirow{4}{*}{\textbf{A}}&CNN-RNN-compressed&{jack white from rock band the dead weather performs during the 44th montreux jazz festival in montreux }\\\cline{2-3}
&Text-summarization&{Little Dragon performing at the House of Blues in Cleveland, OH}\\\cline{2-3}
&Entity-aware(ours)&{singer \textbf{little dragon} performs at the \textbf{house of blues} in \textbf{cleveland}} August 2 2017\\\cline{2-3}
&Human&{\textbf{little dragon} performing at the \textbf{house of blues} in \textbf{cleveland, oh}}\\
\midrule
\multirow{4}{*}{\textbf{B}}&CNN-RNN-compressed&{houston astros starting pitcher brett delivers in the second inning against the cleveland indians at progressive field}\\\cline{2-3}
&Text-summarization&{Red Sox at Orioles 6/2/16}\\\cline{2-3}
&Entity-aware(ours)&{\textbf{baltimore orioles} starting pitcher \textbf{joe kelly} pitches in the first inning against the \textbf{baltimore orioles} at \textbf{baltimore} June 1 2016}\\\cline{2-3}
&Human&{\textbf{joe kelly} of the \textbf{boston red sox} pitches in a game against the \textbf{baltimore orioles} at \textbf{oriole park} at camden yards on june 1, 2016 in \textbf{baltimore, maryland}}\\
\midrule
\multirow{4}{*}{\textbf{C}}&CNN-RNN-compressed&{protestors gesture and hold signs during a protest against what demonstrators call police brutality in mckinney , texas .}\\\cline{2-3}
&Text-summarization&{Toronto, Canada ~ September 9, 2017.}\\\cline{2-3}
&Entity-aware(ours)&{supporters of an ban protest outside the \textbf{toronto international film festival} in \textbf{toronto} September 9 2017}\\\cline{2-3}
&Human&{\textbf{patrick patterson} at the premiere of the \textbf{carter effect}, 2017 \textbf{toronto film festival}}\\
\bottomrule
\end{tabular}
\captionlistentry[table]{Examples of generated entity-aware caption 
.}
\caption{Examples of generated entity-aware caption.}
\label{fig:captione_no_error}
\end{figure*}
\subsection{Flickr Caption Results}
\begin{table*}[ht!]
\small
\centering
\scalebox{1}{
\begin{tabular}
{p{2.8cm}ccccccccc}
\toprule
{}&{BLEU-1}&{BLEU-2}&{BLEU-3}&{BLEU-4}&{METEOR}&{ROUGE}&{CIDEr}&{F1}&{Human$^*$}\\
\midrule
{CNN-RNN-raw}&{7.5}&{2.7}&{0.9}&{0.3}&{2.6}&{7.5}&{1.1}&{1.2}&{0.49}\\
{CNN-RNN-compress}&{8.3}&{3.3}&{1.1}&{0.5}&{3.0}&{9.2}&{1.5}&{2.1}&{0.50}\\
{Text-summarization}&{9.5}&{8.0}&{7.1}&\textbf{6.5}&{9.9}&{11.9}&{17.2}&{35.4}&{0.59}\\
{Entity-aware(ours)}&\textbf{25.5}&\textbf{14.9}&\textbf{8.0}&{{4.7}}&{\textbf{11.0}}&\textbf{21.1}&\textbf{29.9}&{\textbf{39.7}}&\textbf{0.87}\\
\bottomrule
\end{tabular}
}
\caption{Comparison between our entity-aware model and baseline models on various topics. ($^*$ We make human evaluation on 259 images randomly selected.)}
\label{tab:results}
\end{table*}

Table~\ref{tab:results} shows the comparison between our model and the baselines. The scores are much lower than traditional caption generation tasks such as COCO, because we use the real captions as ground-truth. Our model outperforms all the baselines on all metrics except BLEU-4, where Text-summarization model achieves better score. Generally, the model based on textual features (Text-summarization) has better performance than vision-based models (CNN-RNN-raw and CNN-RNN-compressed). It indicates textual summarization algorithm is more effective when it involves specific knowledge generation.
Text-summarization model generates results from documents retrieved by hashtags, so it tends to include some long phrases common in those documents. However the templates generated by our model is based on the language model trained from the news captions, which has different style with Flickr captions. It results in that Text-summarization model achieves better BLEU-4 score. 
Our model improves CIDEr score more significantly compared with other metrics, because CIDEr downweights the n-grams common in all captions, where more specific information such as named entities contribute more to the scores. The experimental results demonstrate that our model is effective to generate image captions with specific knowledge.

\subsection{Analysis}
Figure~\ref{fig:captione_no_error} shows some examples of the captions generated by the entity-aware model. 
\begin{description}
\item \textbf{Good Examples:} (A) in Figure~\ref{fig:captione_no_error} describes the events in the images well (`\textit{performs}') and include correct, well-placed entities, such as `\textit{Little Dragon}' and `\textit{House of Blues}'.
\item \textbf{Relation Error of Filled Entities:} Some of our errors result from ignoring of relations between entities. In Example (B) of Figure~\ref{fig:captione_no_error} our model generate a good template, but connects `\textit{Joe Kelly}', who is actually a pitcher of `\textit{Res Sox}', with `\textit{Baltimore Orioles}' incorrectly. One possible solution is to incorporate relation information when the model fills in the slots with entities.
\item \textbf{Template Error:} Another category of errors results from wrong templates generated by our model. Examples (C) in Figure~\ref{fig:captione_no_error} is about a film festival, but the model generates a template about protest, which is not related to the image. One potential improvement is to incorporate the information from associated tags, such as the number of tags and the named entity types related to the tags, as features during template caption generation to make generated templates dynamically change according to the context.
\end{description}

\section{Related Work}

The goal of image captioning is to automatically generate a natural language sentence given an image. \cite{kulkarni2013babytalk,yang2011corpus,mitchell2012midge} perform object recognition in images and fill hand-made templates with the recognized objects.~\cite{kuznetsova2012collective,kuznetsova2014treetalk} retrieve similar images, parse associated captions into phrases, and compose them into new sentences. Due to the use of static, handmade templates, these approaches are unable to generate a variety of sentence realizations, which can result in poorly generated sentences and requires one to manually create more templates to extend the generation. Our approach overcomes this by dynamically generating the output.

More recent work utilizes neural networks and applies an encoder-decoder model~\cite{cho2014learning}.~\citet{vinyals2015show} use a CNN to encode images into a fixed size vector representation and a LSTM to decode the image representations into a sentence.~\citet{xu2015show} incorporate an attention mechanism~\cite{bahdanau2015neural} and attend to the output from a convolutional layer of a CNN to produce the image representations for the decoder. Instead of encoding a whole image as a vector, \cite{johnson2016densecap} apply R-CNN object detection~\cite{girshick2014rich}, match text snippets to the regions of the image detected by the R-CNN, and use a recurrent neural network (RNN) language model, similar to~\cite{vinyals2015show}, to generate a description of each region.

The surface realization for state-of-the-art neural approaches is impressive, but, in the context of generating entity-aware captions, these methods fall short as they heavily rely on training data for language modeling. \cite{tran2016rich} leverage face and landmark recognition to generate captions containing named persons, but such large-scale training is difficult. Consequently, OOV words like named entities, which are a quintessential aspect of entity-aware captions because OOV words typically represent entities or events, are difficult to generate due to low training probabilities. %or the novel words being out-of-vocabulary. 
Some work has been done to incorporate novel objects into captions~\cite{venugopalan2017captioning}, but this does not address the need to generate entity-aware captions and incorporate contextual information; rather, it gives the ability to generate more fine-grained entities within captions that still lack the necessary context. \cite{feng2013automatic} also generates a caption with named entities, but from associated news articles, in which there is much more textual context than our setting. Our approach uses neural networks to generate dynamic templates and then fills in the templates with specific entities. Thus, we are able to combine the sentence variation and surface realization quality of neural language modeling and the capability to incorporate novel words of template-based approaches.

\section{Conclusions and Future Work}
In this paper we propose a new task which aims to automatically generate entity-aware image descriptions with limited textual information. Experiments on a new benchmark dataset collected from Flickr show that our approach generates more informative captions compared to traditional image captioning methods. Moreover, our two-step approach can easily be applied to other language generation tasks involving specific information. 

In the future, we will expand the entity-aware model to incorporate the relations between candidates when the model fills in the slots, which can avoid the cases such as \textit{`Cristiano Ronaldo of Barcelona'}. We will also make further research on context-aware fine-grained entity typing to train a better template generator. Another research direction based on this work is to develop an end-to-end neural architecture to make the model more flexible without generating a template in the middle.

\section*{Acknowledgments}
This work was supported by the U.S. DARPA AIDA Program No. FA8750-18-2-0014 and U.S. ARL NS-CTA No. W911NF-09-2-0053. The views and conclusions contained in this document are those of the authors and should not be interpreted as representing the official policies, either expressed or implied, of the U.S. Government. The U.S. Government is authorized to reproduce and distribute reprints for Government purposes notwithstanding any copyright notation here on.

%\bibliography{acl2018}
\bibliography{emnlp2018}

\begin{thebibliography}{33}
\expandafter\ifx\csname natexlab\endcsname\relax\def\natexlab#1{#1}\fi

\bibitem[{Auer et~al.(2007)Auer, Bizer, Kobilarov, Lehmann, Cyganiak, and
  Ives}]{auer2007dbpedia}
S{\"o}ren Auer, Christian Bizer, Georgi Kobilarov, Jens Lehmann, Richard
  Cyganiak, and Zachary Ives. 2007.
\newblock Dbpedia: A nucleus for a web of open data.
\newblock In \emph{The semantic web}, pages 722--735. Springer.

\bibitem[{Bahdanau et~al.(2015)Bahdanau, Cho, and Bengio}]{bahdanau2015neural}
Dzmitry Bahdanau, Kyunghyun Cho, and Yoshua Bengio. 2015.
\newblock Neural machine translation by jointly learning to align and
  translate.
\newblock In \emph{Proceedings of the third International Conference on
  Learning Representations}.

\bibitem[{Cho et~al.(2014)Cho, Van~Merri{\"e}nboer, Gulcehre, Bahdanau,
  Bougares, Schwenk, and Bengio}]{cho2014learning}
Kyunghyun Cho, Bart Van~Merri{\"e}nboer, Caglar Gulcehre, Dzmitry Bahdanau,
  Fethi Bougares, Holger Schwenk, and Yoshua Bengio. 2014.
\newblock Learning phrase representations using rnn encoder-decoder for
  statistical machine translation.
\newblock In \emph{Proceedings of the 2014 Conference on Empirical Methods in
  Natural Language Processing}.

\bibitem[{De~Marneffe and Manning(2008)}]{de2008stanford}
Marie-Catherine De~Marneffe and Christopher~D Manning. 2008.
\newblock Stanford typed dependencies manual.
\newblock Technical report, Technical report, Stanford University.

\bibitem[{Deng et~al.(2009)Deng, Dong, Socher, Li, Li, and
  Fei-Fei}]{imagenet_cvpr09}
Jia Deng, Wei Dong, Richard Socher, Li-Jia Li, Kai Li, and Li~Fei-Fei. 2009.
\newblock Imagenet: A large-scale hierarchical image database.
\newblock In \emph{In Proceedings of the 2009 IEEE Conference on Computer
  Vision and Pattern Recognition}.

\bibitem[{Denkowski and Lavie(2014)}]{denkowski2014meteor}
Michael Denkowski and Alon Lavie. 2014.
\newblock Meteor universal: Language specific translation evaluation for any
  target language.
\newblock In \emph{Proceedings of the ninth workshop on statistical machine
  translation.}

\bibitem[{Fang et~al.(2015)Fang, Gupta, Iandola, Srivastava, Deng, Doll{\'a}r,
  Gao, He, Mitchell, Platt et~al.}]{fang2015captions}
Hao Fang, Saurabh Gupta, Forrest Iandola, Rupesh~K Srivastava, Li~Deng, Piotr
  Doll{\'a}r, Jianfeng Gao, Xiaodong He, Margaret Mitchell, John~C Platt,
  et~al. 2015.
\newblock From captions to visual concepts and back.
\newblock In \emph{Proceedings of the 2015 IEEE Conference on Computer Vision
  and Pattern Recognition}.

\bibitem[{Feng and Lapata(2013)}]{feng2013automatic}
Yansong Feng and Mirella Lapata. 2013.
\newblock Automatic caption generation for news images.
\newblock \emph{IEEE transactions on pattern analysis and machine
  intelligence}, 35(4):797--812.

\bibitem[{Girshick et~al.(2014)Girshick, Donahue, Darrell, and
  Malik}]{girshick2014rich}
Ross Girshick, Jeff Donahue, Trevor Darrell, and Jitendra Malik. 2014.
\newblock Rich feature hierarchies for accurate object detection and semantic
  segmentation.
\newblock In \emph{Proceedings of the 2014 IEEE conference on computer vision
  and pattern recognition}.

\bibitem[{He et~al.(2016)He, Zhang, Ren, and Sun}]{he2016deep}
Kaiming He, Xiangyu Zhang, Shaoqing Ren, and Jian Sun. 2016.
\newblock Deep residual learning for image recognition.
\newblock In \emph{Proceedings of the IEEE conference on computer vision and
  pattern recognition}, pages 770--778.

\bibitem[{Hochreiter and Schmidhuber(1997)}]{hochreiter1997long}
Sepp Hochreiter and J{\"u}rgen Schmidhuber. 1997.
\newblock Long short-term memory.
\newblock \emph{Neural computation}, 9(8):1735--1780.

\bibitem[{Johnson et~al.(2016)Johnson, Karpathy, and
  Fei-Fei}]{johnson2016densecap}
Justin Johnson, Andrej Karpathy, and Li~Fei-Fei. 2016.
\newblock Densecap: Fully convolutional localization networks for dense
  captioning.
\newblock In \emph{Proceedings of the IEEE Conference on Computer Vision and
  Pattern Recognition}, pages 4565--4574.

\bibitem[{Karpathy and Fei-Fei(2015)}]{karpathy2015deep}
Andrej Karpathy and Li~Fei-Fei. 2015.
\newblock Deep visual-semantic alignments for generating image descriptions.
\newblock In \emph{Proceedings of the 2015 IEEE Conference on Computer Vision
  and Pattern Recognition}.

\bibitem[{Kulkarni et~al.(2013)Kulkarni, Premraj, Ordonez, Dhar, Li, Choi,
  Berg, and Berg}]{kulkarni2013babytalk}
Girish Kulkarni, Visruth Premraj, Vicente Ordonez, Sagnik Dhar, Siming Li,
  Yejin Choi, Alexander~C Berg, and Tamara~L Berg. 2013.
\newblock Babytalk: Understanding and generating simple image descriptions.
\newblock \emph{IEEE Transactions on Pattern Analysis and Machine
  Intelligence}, 35(12):2891--2903.

\bibitem[{Kuznetsova et~al.(2012)Kuznetsova, Ordonez, Berg, Berg, and
  Choi}]{kuznetsova2012collective}
Polina Kuznetsova, Vicente Ordonez, Alexander~C Berg, Tamara~L Berg, and Yejin
  Choi. 2012.
\newblock Collective generation of natural image descriptions.
\newblock In \emph{Proceedings of the 50th Annual Meeting of the Association
  for Computational Linguistics}.

\bibitem[{Kuznetsova et~al.(2014)Kuznetsova, Ordonez, Berg, and
  Choi}]{kuznetsova2014treetalk}
Polina Kuznetsova, Vicente Ordonez, Tamara~L Berg, and Yejin Choi. 2014.
\newblock Treetalk: Composition and compression of trees for image
  descriptions.
\newblock \emph{Transactions of the Association of Computational Linguistics}.

\bibitem[{Lin(2004)}]{lin2004rouge}
Chin-Yew Lin. 2004.
\newblock Rouge: A package for automatic evaluation of summaries.
\newblock \emph{Text Summarization Branches Out}.

\bibitem[{Lin et~al.(2014)Lin, Maire, Belongie, Hays, Perona, Ramanan,
  Doll{\'a}r, and Zitnick}]{lin2014microsoft}
Tsung-Yi Lin, Michael Maire, Serge Belongie, James Hays, Pietro Perona, Deva
  Ramanan, Piotr Doll{\'a}r, and C~Lawrence Zitnick. 2014.
\newblock Microsoft coco: Common objects in context.
\newblock In \emph{Proceedings of the 2014 European Conference on Computer
  Vision}.

\bibitem[{Manning et~al.(2014)Manning, Surdeanu, Bauer, Finkel, Bethard, and
  McClosky}]{manning2014stanford}
Christopher Manning, Mihai Surdeanu, John Bauer, Jenny Finkel, Steven Bethard,
  and David McClosky. 2014.
\newblock The stanford corenlp natural language processing toolkit.
\newblock In \emph{Proceedings of 52nd annual meeting of the association for
  computational linguistics: system demonstrations}, pages 55--60.

\bibitem[{Mitchell et~al.(2012)Mitchell, Han, Dodge, Mensch, Goyal, Berg,
  Yamaguchi, Berg, Stratos, and Daum{\'e}~III}]{mitchell2012midge}
Margaret Mitchell, Xufeng Han, Jesse Dodge, Alyssa Mensch, Amit Goyal, Alex
  Berg, Kota Yamaguchi, Tamara Berg, Karl Stratos, and Hal Daum{\'e}~III. 2012.
\newblock Midge: Generating image descriptions from computer vision detections.
\newblock In \emph{Proceedings of the 13th Conference of the European Chapter
  of the Association for Computational Linguistics}.

\bibitem[{Pan et~al.(2015)Pan, Cassidy, Hermjakob, Ji, and Knight}]{Pan2015}
Xiaoman Pan, Taylor Cassidy, Ulf Hermjakob, Heng Ji, and Kevin Knight. 2015.
\newblock Unsupervised entity linking with abstract meaning representation.
\newblock In \emph{Proceedings of the 2015 Conference of the North American
  Chapter of the Association for Computational Linguistics--Human Language
  Technologies}.

\bibitem[{Papineni et~al.(2002)Papineni, Roukos, Ward, and
  Zhu}]{papineni2002bleu}
Kishore Papineni, Salim Roukos, Todd Ward, and Wei-Jing Zhu. 2002.
\newblock Bleu: a method for automatic evaluation of machine translation.
\newblock In \emph{Proceedings of the 40th annual meeting on association for
  computational linguistics}.

\bibitem[{Rashtchian et~al.(2010)Rashtchian, Young, Hodosh, and
  Hockenmaier}]{rashtchian2010collecting}
Cyrus Rashtchian, Peter Young, Micah Hodosh, and Julia Hockenmaier. 2010.
\newblock Collecting image annotations using amazon's mechanical turk.
\newblock In \emph{Proceedings of the 2010 Conference of the North American
  Chapter of the Association for Computational Linguistics--Human Language
  Technologies Workshop on Creating Speech and Language Data with Amazon's
  Mechanical Turk}.

\bibitem[{Tran et~al.(2016)Tran, He, Zhang, Sun, Carapcea, Thrasher, Buehler,
  and Sienkiewicz}]{tran2016rich}
Kenneth Tran, Xiaodong He, Lei Zhang, Jian Sun, Cornelia Carapcea, Chris
  Thrasher, Chris Buehler, and Chris Sienkiewicz. 2016.
\newblock Rich image captioning in the wild.
\newblock In \emph{Proceedings of the 2016 IEEE Conference on Computer Vision
  and Pattern Recognition Workshops}.

\bibitem[{Tran and Nguyen(2017)}]{tran2017nlg}
Van-Khanh Tran and Le-Minh Nguyen. 2017.
\newblock Natural language generation for spoken dialogue system using rnn
  encoder-decoder networks.
\newblock In \emph{Proceedings of the 21st Conference on Computational Natural
  Language Learning}.

\bibitem[{Vanderwende et~al.(2007)Vanderwende, Suzuki, Brockett, and
  Nenkova}]{vanderwende2007beyond}
Lucy Vanderwende, Hisami Suzuki, Chris Brockett, and Ani Nenkova. 2007.
\newblock Beyond sumbasic: Task-focused summarization with sentence
  simplification and lexical expansion.
\newblock \emph{Information Processing \& Management}, 43(6):1606--1618.

\bibitem[{Vedantam et~al.(2015)Vedantam, Lawrence~Zitnick, and
  Parikh}]{vedantam2015cider}
Ramakrishna Vedantam, C~Lawrence~Zitnick, and Devi Parikh. 2015.
\newblock Cider: Consensus-based image description evaluation.
\newblock In \emph{Proceedings of the IEEE conference on computer vision and
  pattern recognition}, pages 4566--4575.

\bibitem[{Venugopalan et~al.(2017)Venugopalan, Hendricks, Rohrbach, Mooney,
  Darrell, and Saenko}]{venugopalan2017captioning}
Subhashini Venugopalan, Lisa~Anne Hendricks, Marcus Rohrbach, Raymond Mooney,
  Trevor Darrell, and Kate Saenko. 2017.
\newblock Captioning images with diverse objects.
\newblock In \emph{Proceedings of the 2017 IEEE Conference on Computer Vision
  and Pattern Recognition}.

\bibitem[{Vinyals et~al.(2015)Vinyals, Toshev, Bengio, and
  Erhan}]{vinyals2015show}
Oriol Vinyals, Alexander Toshev, Samy Bengio, and Dumitru Erhan. 2015.
\newblock Show and tell: A neural image caption generator.
\newblock In \emph{Proceedings of the 2015 IEEE Conference on Computer Vision
  and Pattern Recognition}.

\bibitem[{Wang et~al.(2015)Wang, Zheng, Ma, Fox, and Ji}]{wang2015}
Han Wang, Jin~Guang Zheng, Xiaogang Ma, Peter Fox, and Heng Ji. 2015.
\newblock Language and domain independent entity linking with quantified
  collective validation.
\newblock In \emph{Proceedings of the 2015 Conference on Empirical Methods in
  Natural Language Processing}.

\bibitem[{Wen et~al.(2015)Wen, Gasic, Mrksic, Su, Vandyke, and
  Young}]{wen2015semantically}
Tsung-Hsien Wen, Milica Gasic, Nikola Mrksic, Pei-Hao Su, David Vandyke, and
  Steve Young. 2015.
\newblock Semantically conditioned lstm-based natural language generation for
  spoken dialogue systems.
\newblock In \emph{Proceedings of the 2015 Conference on Empirical Methods in
  Natural Language Processing}.

\bibitem[{Xu et~al.(2015)Xu, Ba, Kiros, Cho, Courville, Salakhutdinov, Zemel,
  and Bengio}]{xu2015show}
Kelvin Xu, Jimmy Ba, Ryan Kiros, Kyunghyun Cho, Aaron~C Courville, Ruslan
  Salakhutdinov, Richard~S Zemel, and Yoshua Bengio. 2015.
\newblock Show, attend and tell: Neural image caption generation with visual
  attention.
\newblock In \emph{Proceedings of the 2015 International Conference on Machine
  Learning}.

\bibitem[{Yang et~al.(2011)Yang, Teo, Daum{\'e}~III, and
  Aloimonos}]{yang2011corpus}
Yezhou Yang, Ching~Lik Teo, Hal Daum{\'e}~III, and Yiannis Aloimonos. 2011.
\newblock Corpus-guided sentence generation of natural images.
\newblock In \emph{Proceedings of the Conference on Empirical Methods in
  Natural Language Processing}.

\end{thebibliography}
\bibliographystyle{acl_natbib_nourl}

% \appendix
% \appendix

% \section{Entity-aware Image Caption Examples}
% \label{sec:appendix}
% \appendix
% \section{Examples of generated Template based on coarse-grained and fine-grained entity types.}
% \section{Examples of Entity-aware Image Captions generated using Flickr Data}
\end{document}